\documentclass[letterpaper, 10 pt, conference]{ieeeconf}  

\IEEEoverridecommandlockouts                              

\usepackage{graphicx} 
\usepackage{mathtools} 
\usepackage{amssymb}  
\usepackage{bm}
\usepackage{xcolor}
\usepackage{multirow,tabularx}
\usepackage{subcaption}
\usepackage{caption}
\usepackage{hyperref}
\hypersetup{
    colorlinks=true,
    linkcolor=black,
    filecolor=black,      
    urlcolor=blue,
}

\title{\LARGE \bf
Desperate Times Call for Desperate Measures: \\ Towards Risk-Adaptive Task Allocation
}

\author{Max Rudolph$^{\dagger}$, Sonia Chernova$^{\dagger}$, Harish Ravichandar$^{\dagger}$
\thanks{*This work was supported by the Army Research Lab under Grant W911NF-17-2-0181 (DCIST CRA) and W911NF-20-2-0036.}
\thanks{$^{\dagger}$Georgia Institute of Technology {\tt\small \{maxrudolph, chernova, harish.ravichandar\}@gatech.edu}}%
}

\begin{document}

\maketitle
\thispagestyle{empty}
\pagestyle{empty}

\begin{abstract}
Multi-robot task allocation (MRTA) problems involve optimizing the allocation of robots to tasks. MRTA problems are known to be challenging when tasks require multiple robots and the team is composed of heterogeneous robots. These challenges are further exacerbated when we need to account for uncertainties encountered in the real-world. In this work, we address coalition formation in heterogeneous multi-robot teams with uncertain capabilities. We specifically focus on tasks that require coalitions to collectively satisfy certain minimum requirements. Existing approaches to uncertainty-aware task allocation either maximize expected pay-off (risk-neutral approaches) or improve worst-case or near-worst-case outcomes (risk-averse approaches). Within the context of our problem, we demonstrate the inherent limitations of unilaterally ignoring or avoiding risk and show that these approaches can in fact reduce the probability of satisfying task requirements. Inspired by models that explain foraging behaviors in animals, we develop a \textit{risk-adaptive} approach to task allocation. Our approach adaptively switches between risk-averse and risk-seeking behavior in order to maximize the probability of satisfying task requirements. Comprehensive numerical experiments conclusively demonstrate that our risk-adaptive approach outperforms risk-neutral and risk-averse approaches. We also demonstrate the effectiveness of our approach using a simulated multi-robot emergency response scenario.
\end{abstract}

\section{Introduction}



Multi-robot systems have to deal with various sources of uncertainties when operating in the real-world. As such, we require models and approaches that account for such uncertainty when coordinating a team of robots. This need has inspired a considerable amount of recent efforts aimed at developing risk-aware approaches to multi-robot coordination that explicitly account for different forms of uncertainty (see \cite{zhou_multi-robot_2021} for a comprehensive survey). Indeed, such risk-aware approaches have been shown to be significantly more successful than approaches that ignore uncertainty.

In this work, we address multi-robot task allocation (MRTA) problems that involve uncertainty. Within MRTA, we focus on the single-task robots multi-robot tasks instantaneous assignment (ST-MR-IA) problem in heterogeneous multi-robot teams (see \cite{gerkey2004,korsah2013comprehensive} for detailed treatments of the various categories of MRTA). The ST-MR-IA problem is also referred to as the coalition formation problem. While there are various sources of uncertainty, we focus on the uncertainty in robots' capabilities. Such uncertainties arise either due to potential failures or due to modeling large teams of robots into a small number of groups (e.g., \cite{prorok2017impact,ravichandar2020}).

Existing approaches to risk-based task allocation fall into one of two categories. First, \textit{risk-neutral} approaches focus on the expected value of pay-off or cost (e.g., \cite{prorok_redundant_2019,choudhury2020dynamic}). Second, \textit{risk-averse} approaches avoid worst-case and near-worst case outcomes (e.g., \cite{nam_analyzing_2017,ravichandar2020}).


In this work, we argue that neither ignoring nor avoiding risk might be sufficient for a certain class of task allocation problems. Specifically, we focus on task allocation problems that require each coalition to satisfy certain minimum capability-based requirements associated with the assigned task. Examples of such minimum requirements involve capabilities such as collective payload, fuel level, and specialized equipment. As such, falling short of these requirements would result in categorical task failure. In such scenarios, we show that it might be necessary to resort to riskier solutions when faced with dire circumstances.

Our view of risk management is inspired by a rich body of work on risk-sensitive foraging behavior in animals (e.g., \cite{caraco_empirical_1980}). This literature demonstrated that animals will prefer to forage under safer conditions (with low-variance on available food) if they are able to meet their calorific needs. However, if such safe sources of food fail to meet their energy demands, they would resort to risk-prone foraging strategies with costlier worst-case outcomes. Indeed, it was shown that this \textit{adaptive} behavior is optimal in the sense that it minimizes the probability of starvation~\cite{stephens_logic_1981}.

Inspired by adaptive animal behavior, we formalize and develop a \textit{risk-adaptive} approach to task allocation and coalition formation. Our approach is capable of autonomously choosing between safer and riskier options. In contrast to maximizing a stochastic pay-off, our approach solves a constrained optimization problem to explicitly optimize the probability of meeting or surpassing minimum requirements. 


We evaluate our approach using detailed numerical evaluations and simulated robot experiments on the \emph{Robotarium}~\cite{wilson2020} simulator. In each of the experiments, we compared our risk-adaptive approach against three baselines: random, risk-neutral, and risk-averse allocation approaches. The results conclusively demonstrate the benefits of a risk-adaptive approach over the baselines in terms of task success rates.

In summary, our core contributions include:
\begin{itemize}
    \item A formalism for risk-based task allocation that acknowledges the benefit of risk-seeking behavior when safer options are unlikely to satisfy minimum requirements.
    \item A risk-adaptive task allocation algorithm that autonomously switches between risk-seeking and risk-averse behavior to better satisfy task requirements.
\end{itemize}

\subsection*{Related work}

MRTA problems are typically categorized based on three dimensions: i) single-task (ST) robots vs. multi-task (MR) robots, ii) single-robot (SR) tasks vs. multi-robot (MR) tasks, and iii) instantaneous assignment (IA) vs. time-extended assignment (TA)~\cite{gerkey2004,korsah2013comprehensive}. Our work falls under the category of ST-MR-IA (also called coalition formation) and is known to be NP-hard. While there is a large body of work associated with the various categories of MRTA, we limit our discussion to approaches focused on coalition formation.

Coalition formation has been tackled by a wide variety of approaches. Notable examples include auction-based methods that rely on effective biding mechanisms (e.g., \cite{guerrero2003multi,xie2018mutual}), utility-based methods that attempt to jointly maximize the total utility (e.g., \cite{vig2006multi,parker2006building}), and the more-recent trait-based approaches that attempt to satisfy trait requirements associated with each task. Our approach falls under the category of trait-based methods, which do not assume knowledge of the utility of assigning each robot or coalition to each task. Instead, we allow for task requirements to be specified in terms of the capabilities necessary to perform the task. 


The methods discussed so far do not account for the various sources of uncertainty that a multi-robot system might face in the real-world. Recent attempts have focused on explicitly accounting for such uncertainty~\cite{zhou_multi-robot_2021}. Existing approaches to risk-based task allocation are either risk-neutral or risk-averse. Risk-neutral approaches focus on the expected value of pay-off or cost~\cite{prorok_redundant_2019,choudhury2020dynamic}. Risk-averse approaches take variance into account and try to avoid worst-case, potentially leading to highly conservative outcomes. 

Recent work has demonstrated that risk-averse methods can be made less conservative by considering more nuanced measures of risk (e.g., mean-variance~\cite{yang_algorithm_2018,ravichandar2020}, and conditional value at risk (CVaR)~\cite{nam_analyzing_2017,sharma2020risk}) that allow for a user-specified level of risk. However, these methods require that the user \textit{predetermines} the desired risk tolerance (e.g., regularizer $\lambda$ in mean-variance optimization and risk parameter $\alpha$ in VaR or CVaR). This explicit and \textit{a priori} specification of risk tolerance places these approaches on a \textit{static} point on the spectrum from risk-averse to risk-seeking, irrespective of the current context. In contrast, our approach \textit{adaptively} determines where to fall on this spectrum depending on the context as determined by task requirements and the availability of resources. As such, our approach adapts to the particulars of the problem, producing riskier or more conservative allocations depending on what will maximize the probability of task success. Further, unlike most existing approaches that optimize a single-dimensional pay-off variable, we can handle multi-dimensional requirements.


Finally, note that most existing risk-aware task allocation approaches are limited to single-robot tasks~\cite{nam_analyzing_2017,choudhury2020dynamic}, homogeneous agents~\cite{prorok_redundant_2019}, or both~\cite{yang_algorithm_2017,sharma2020risk}. To the best of our knowledge, our work represents the first attempt to solve risk-aware coalition formation in heterogeneous teams.


\section{Modeling Framework}

To provide context, we first introduce our basic modeling principles, which are adapted from our prior work~\cite{ravichandar2020}. 

\subsection{Species}
Consider a team of $N$ heterogeneous robots. We take a group modeling approach~\cite{albrecht2018autonomous} and model the team of robots as being made of $S$ species (i.e. robot types). Examples of such species include a group of UAVs and a group of ground vehicles. By utilizing such an aggregate model at the level of robot types, we gain computational efficiency over alternative approaches that model each robot individually. 

\subsection{Traits}
When modeling traits (i.e., capabilities), we take into account the fact that robots within a particular species may not share identical traits. For instance, not all UAVs will share the same speed or carrying capacity. As such, we model the traits of the $s$th species as $q^{(s)} \sim \mathcal{N}(\mu_{q^{(s)}}, \Sigma_{q^{(s)}})$, where $\mu_{q^{(s)}} \in \mathbb{R}_+^{U}$ and $\Sigma_{q^{(s)}} \in \mathbb{R}_+^{U \times U}$ are the expected trait vector and the corresponding diagonal covariance matrix indicating that each trait of the $s$th species is an independent Gaussian random variable. 
Taken together, the traits of the entire team are denoted by the \emph{stochastic species-trait matrix} $\bm{Q}$ with $\bm{\mu}_Q \in \mathbb{R}^{S \times U}_{+}$ containing the expected values. Specifically the $su$th element of $\bm{\mu}_Q$ denotes the expected value of the $u$th trait of the $s$th species.
Similarly, the variances associated with each trait of each species is contained in the matrix $\bm{\mathrm{Var}}_Q \in \mathbb{R}^{S \times U}_{+}$. The $su$th element of $\bm{\mathrm{Var}}_Q$ denotes the variance of the $u$th trait of the $s$th species. 

\subsection{Tasks}

Let the team be tasked with solving $M$ concurrent tasks, each with its own set of trait requirements denoted by $Y^*_m \in \mathbb{R}_+^U, \forall m=1,2,\cdots,M$. To successfully complete the tasks, the team has to form coalitions such that each coalition collectively meets or surpasses the corresponding task's trait requirements. The trait requirements for all the tasks can be represented by a \emph{task requirements matrix} $\bm{Y}^* \in \mathbb{R}_+^{M \times U}$.

\subsection{Agent Assignment}

The assignment of agents from species $s$ across the $M$ tasks is denoted by $\mathrm{x}^{(s)} =$ $[x^{(s)}_{1}, x^{(s)}_{2},$ $\cdots x^{(s)}_{M}]^T$ $\in \mathbb{N}^M$. Thus, the assignment of the whole team across the tasks can be described using the \emph{assignment matrix} $\bm{X} = [\mathrm{x}^{(1)}, \mathrm{x}^{(2)}, \cdots, \mathrm{x}^{(S)}] \in \mathbb{N}^{M \times S}$.

\subsection{Trait Aggregation}
Finally, the aggregation of various traits assigned across all the tasks is denoted by the \emph{stochastic trait distribution matrix} $\bm{Y} \in \mathbb{R}_{+}^{M \times U}$, and can be computed as
\begin{equation}
    \bm{Y} = \bm{X}\bm{Q}
\end{equation}
Note that $\bm{Y}=[Y_1,\cdots,Y_M]^T$ is composed of $M$ Gaussian random variables (one for each task) due to the fact that $\bm{Q}$ is composed of $S$ Gaussian random variable (one for each species). Thus, the expected value of $\bm{Y}$ is given by
\begin{equation}
    \bm{\mu}_Y =  \bm{X} \bm{\mu}_Q \label{eq:mean_trait_dist_tasks}
\end{equation}
and the variance of each element of $\bm{Y}$ given by
\begin{equation}
    \bm{\mathrm{Var}}_{Y} = (\bm{X} \odot \bm{X})\ \bm{\mathrm{Var}}_Q \label{eq:var_trait_dist_tasks}
\end{equation}
where $\odot$ denotes the Hadamard (element-wise) product. 

\section{Risk-Adaptive Task Allocation}

In this section, we introduce the notion of risk-adaptive task allocation. We begin by considering the trait requirements associated with all the tasks. Let the minimum trait requirements associated with the $m$th task be given by $Y_m^*\in \mathbb{R}_+^{U}$. Thus, the probability of successfully performing the $m$th task is given by 
\begin{equation}
    P_m(\text{success}) = P(Y_m \succ Y^*_m) \label{eq:success_prob}
\end{equation}
where $\succ$ denotes the element-wise grater than operator. Thus, the success of each task is given by a multi-variate normal cumulative density function.

\begin{figure}
    \centering
    \includegraphics[width=\columnwidth]{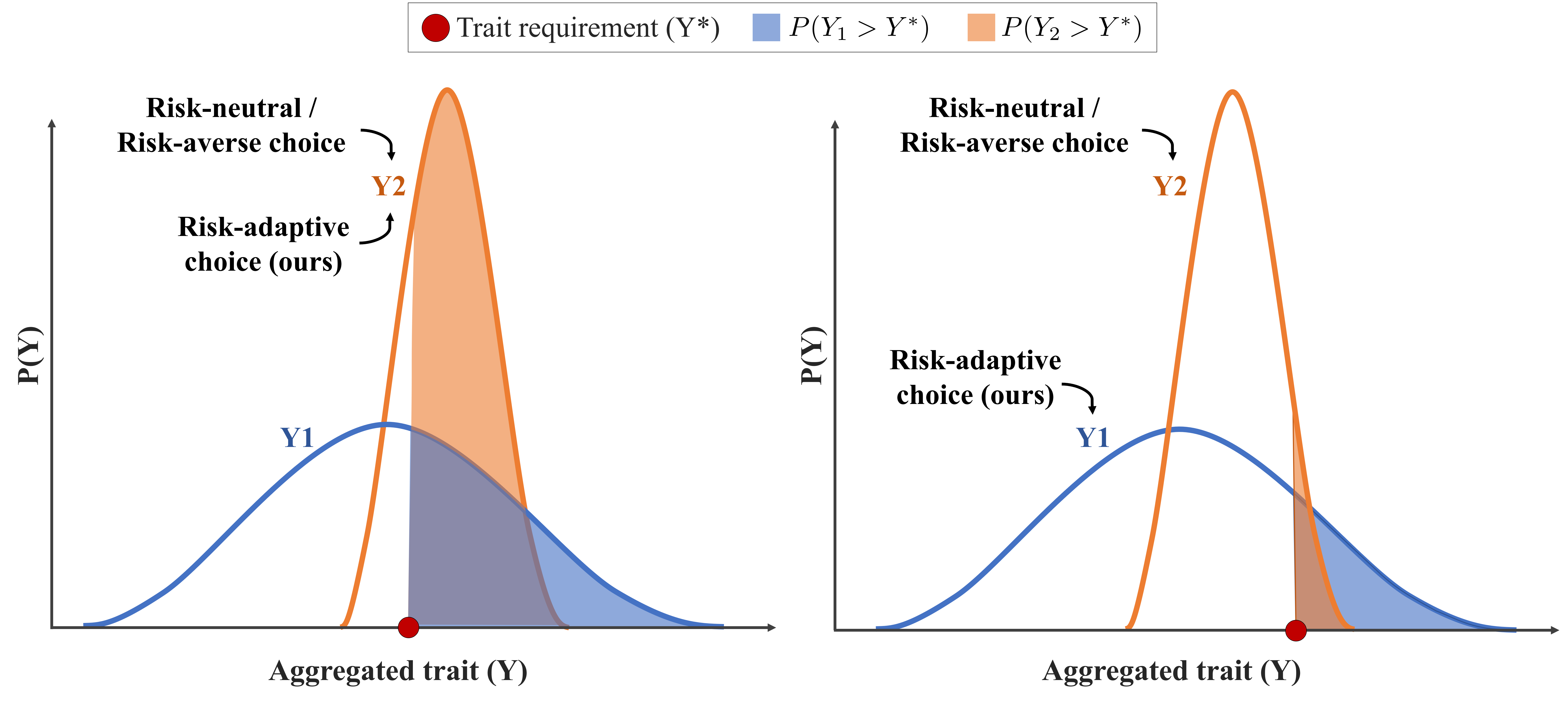}
    \caption{Consider two scenarios with different choices for coalitions that will result in different stochastic aggregation of capabilities (blue and orange curves). In each scenario, the minimum trait requirement in each scenarios is depicted by a red circle, and the area of each shaded region denotes the corresponding probability of satisfying the task requirements. 
    }
    \label{fig:illustration}
\end{figure}

\subsection{Illustrative Example}
\label{subsec:example}

To illustrate the benefits of a risk-adaptive approach, let us consider an example task that requires a coalition of robots that can collectively satisfy a single-dimensional trait requirement, such as payload or fuel. Without loss of generality, let us analyze two options for the coalition with different aggregate traits ($Y_1$ and $Y_2$). Indeed, given the probabilistic nature of our capabilities model, the aggregate trait of each coalition represents a probability distribution. When a safer (orange $Y2$) option exists that can satisfy the trait requirement $Y^*$ in expectation (as in Fig. \ref{fig:illustration}, left), our risk-adaptive approach would prefer it, behaving similarly to risk-neutral or risk-averse approaches. In contrast, when neither coalition can satisfy the trait requirement in expectation (as in Fig. \ref{fig:illustration}, right), our approach would adaptively choose the riskier option (blue $Y1$), as it maximizes the chances of satisfying the minimum requirement.

\subsection{Rationale}
\label{sec:rationale}

The analysis of animal foraging behavior in \cite{stephens_logic_1981} can be easily extended to explain why a risk-adaptive strategy improves the probability of success in (\ref{eq:success_prob}).

Consider a potential allocation such that the expected value of the resulting trait aggregation satisfies the desired trait requirements (i.e., $\bm{\mu}_Y \succ \bm{Y}^*$). Under this circumstance, it is clear that the probability of success (i.e., $P(\bm{Y} \succ \bm{Y}^*)$) will increase only if the variance (i.e., $\bm{\mathrm{Var}}_{Y}$) decreases. Thus, our risk-adaptive approach operates in a \textit{risk-averse regime} when $\bm{\mu}_Y \succ \bm{Y}^*$ as it prefers allocations with smaller variances if their expected values are similar. This observation further explains the choices in Fig. \ref{fig:illustration} (left).

Similarly, consider a potential allocation such that the expected value of the trait aggregation fails to satisfy the desired trait requirements (i.e., $\bm{\mu}_Y \prec \bm{Y}^*$). Under this circumstance, it is clear that the probability of success (i.e., $P(\bm{Y} \succ \bm{Y}^*)$) will increase only if the variance (i.e., $\bm{\mathrm{Var}}_{Y}$) increases. Thus, our risk-adaptive approach operates in a \textit{risk-seeking regime} when $\bm{\mu}_Y \prec \bm{Y}^*$ as it prefers allocations with larger variances if their expected values are similar. In contrast, risk-averse approaches will continue to prefer smaller variances as they optimize for worst-case outcomes. However, as a result, risk-averse approaches will inadvertently decrease the probability of success. This observation further explains the choices in Fig. \ref{fig:illustration} (right).

\subsection{Constrained Optimization}
\label{subsec:optimization}

Given the model for task success, we turn to the problem of optimizing the probability of success. Note that the example from \ref{subsec:example} is focused on a single task. Our problem consists of forming coalitions for $M$ tasks when provided a fixed number of agents from each species. Thus, we simultaneously optimize the chances of satisfying the requirements for all tasks.

We cast our risk-adaptive task allocation problem in the form of the following max-min optimization problem
\begin{align}
    \bm{X}^* = \arg\ \max_{\bm{X}} &\ \min_{m} \, \log P(Y_m \succ Y^*_m) \label{eq:propo_cost_min} \\
    \mathrm{s.t.} &\ \bm{X}^T \cdot \bm{1} \leq N_A \label{eq:robot_limit} \\
    &\ \bm{X} \in \mathbb{Z}_+^{M \times S} \label{eq:integer_const}
\end{align}
where $N_A \in \mathbb{Z}_+^S$ is a vector of the number of agents in each species. An alternative strategy would be to replace the objective function in (\ref{eq:propo_cost_min}) with the average or sum of individual task probabilities. However, such an objective function will not discourage disproportionately different success probabilities across tasks, resulting in skewed allocation of robots to tasks and unintended prioritization.

Note that the optimization problem in (\ref{eq:propo_cost_min})-(\ref{eq:integer_const}) represents a considerably challenging nonlinear constrained integer program. In this work, we approximately solve this problem by relaxing the integer constraint in (\ref{eq:integer_const}) and replacing it with the constraint $\bm{X} \in \mathbb{R}_+^{M \times S}$. Further, given the non-convex natural of the objective function, we employ a global optimization technique that performs a scatter search to provide multiple initial conditions for a local nonlinear program solver~\cite{ugray2007scatter}. Finally, we convert every element of the optimized assignment matrix $\bm{X}^*$ into an integer while ensuring that the constraint in (\ref{eq:robot_limit}) is satisfied.

In practice, we initialize the allocation matrix $\bm{X}$ using a risk-neutral solution. As such, the global optimization attempts to improve the probability of success when possible by resorting to riskier options when appropriate. If safer options exists, our approach will choose allocations that are similar to that of risk-neutral or risk-averse approach.

\section{Experiments}

We evaluated our approach with two experiments: 1) a numerical simulation using teams of varying size, trait distribution, and task requirement, and 2) a simulated robot experiment in the \emph{Robotarium} multi-robot testbed simulator~\cite{wilson2020} that samples robots from a given trait distribution in order to complete two example tasks. Across all experiments, we used MATLAB's \verb|GlobalSearch| function to approximately solve the optimization problem in (\ref{eq:propo_cost_min})-(\ref{eq:integer_const}) as detailed in Section \ref{subsec:optimization} and the baselines. All experiments were conducted using a 2.6 GHz 6-core Intel i7 Processor\footnote{Source code available \href{https://github.com/maxrudolph1/risk_adaptive_task_allocation}{here}.}. On average, the optimization took approximately 0.4 seconds for each baseline and 3.0 seconds for our method. The difference in computation time is due to the fact that, unlike the baselines, our approach solves a non-convex problem.


\subsection{Baselines}
\label{sec:opt_crit}

In all of our experiments, we compared the performance of our method with that of the following three baselines:
\vspace{3pt}

\noindent \textit{1. Random} baseline uniformly randomly allocates the available agents to all the tasks.

\vspace{3pt}
\noindent \textit{2. Risk-neutral} baseline allocates agents such that the expected trait aggregation satisfies the trait requirements. This baseline is similar in spirit to existing approaches that focus on expected pay-off (e.g., \cite{prorok_redundant_2019,choudhury2020dynamic}). To this end, it solves the following optimization problem
\begin{align}
    \bm{X}^* = \arg \min_{\bm{X}} &\ ||\max (Y^* - \bm{X}\mu_Q , 0)||^2_\text{F} \nonumber \\
    \mathrm{s.t.} &\ \bm{X}^T \cdot \bm{1} \leq N_A \nonumber \\
    &\ \bm{X} \in \mathbb{Z}_+^{M \times S} \nonumber
\end{align}
where $|| \cdot ||_\text{F}$ denotes the Frobenius norm.

\vspace{3pt}
\noindent \textit{3. Risk-averse} baseline allocates agents such that worst-case or near worst-case outcomes are avoided. This baseline is similar in spirit to existing approaches that rely on mean-variance optimization (e.g. \cite{ravichandar2020,yang_algorithm_2018}) as it solves the following optimization problem
\begin{align}
    \bm{X}^* = \arg \min_{\bm{X}} &\ ||\max (Y^* - \bm{X}\mu_Q , 0)||^2_\text{F} + \lambda ||\mathrm{Var}_{Y} ||^2_\text{F} \nonumber \\
    \mathrm{s.t.} &\ \bm{X}^T \cdot \bm{1} \leq N_A \nonumber \\
    &\ \bm{X} \in \mathbb{Z}_+^{M \times S}
    \nonumber
\end{align}
where $\lambda \in \mathbb{R}_+$ is a regularization coefficient.

Similar to our proposed risk-adaptive approach, the optimization problems associated with both the risk-neutral and risk-averse baselines were solved approximately by relaxing the integer constraint and utilizing MATLAB's \verb|GlobalSearch| function to ensure fair comparisons. Further, we utilized sequential quadratic programming (SQP) as the local solver in global optimization for all algorithms with the maximum number of iterations set to $10,000$.

\subsection{Numerical Simulations}

We first analyzed the performance of our method and that of the baselines using numerical simulations. To this end, we simulated $100$ independent coalition formation problems involving $S=3$ species each, $U=3$ traits, and $M=3$ tasks. To generate a heterogeneous teams, we ensured that each species had a dominant trait (i.e., higher expected trait value than its other traits). Simulation of such dominant traits is motivated by the fact that real-world robots are often optimized for a few attributes while trading-off others (e.g., speed vs. payload). On average, the variance of the dominant trait is smaller than that of the non-dominant traits. 
During each simulation run, parameters of the robot trait distribution ($\bm{\mu}_Q$ and $\bm{\mathrm{Var}}_Q$), number of robots per species ($N_A$), and task trait requirement ($Y^*$) are uniformly randomly sampled from ranges described in Table \ref{tab:trait_table}. 

\begin{table}[ht]
    \centering
    \begin{tabular}{|c|c|}
        \hline
        Parameter & Distribution Range \\
        \hline
        Dominant trait $\mu$ & $\mathcal{U}(4,5)$ \\
        Non-dominant trait $\mu$ & $\mathcal{U}(0,1)$ \\
        Dominant Trait $\mathrm{Var}$ & $\mathcal{U}(0,0.5)$ \\
        Non-Dominant Trait $\mathrm{Var}$  & $\mathcal{U}(0,1)$ \\ 
        \# robots per species ($N_A$)  & $\mathcal{U}(\{5,\dots,15\})$ \\
        \hline
    \end{tabular}
    \caption{Design parameter sampling ranges}
    \label{tab:trait_table}
\end{table}

\begin{figure}[ht]
    \centering
    \includegraphics[trim=10 10 10 3,clip, width=1.0\columnwidth]{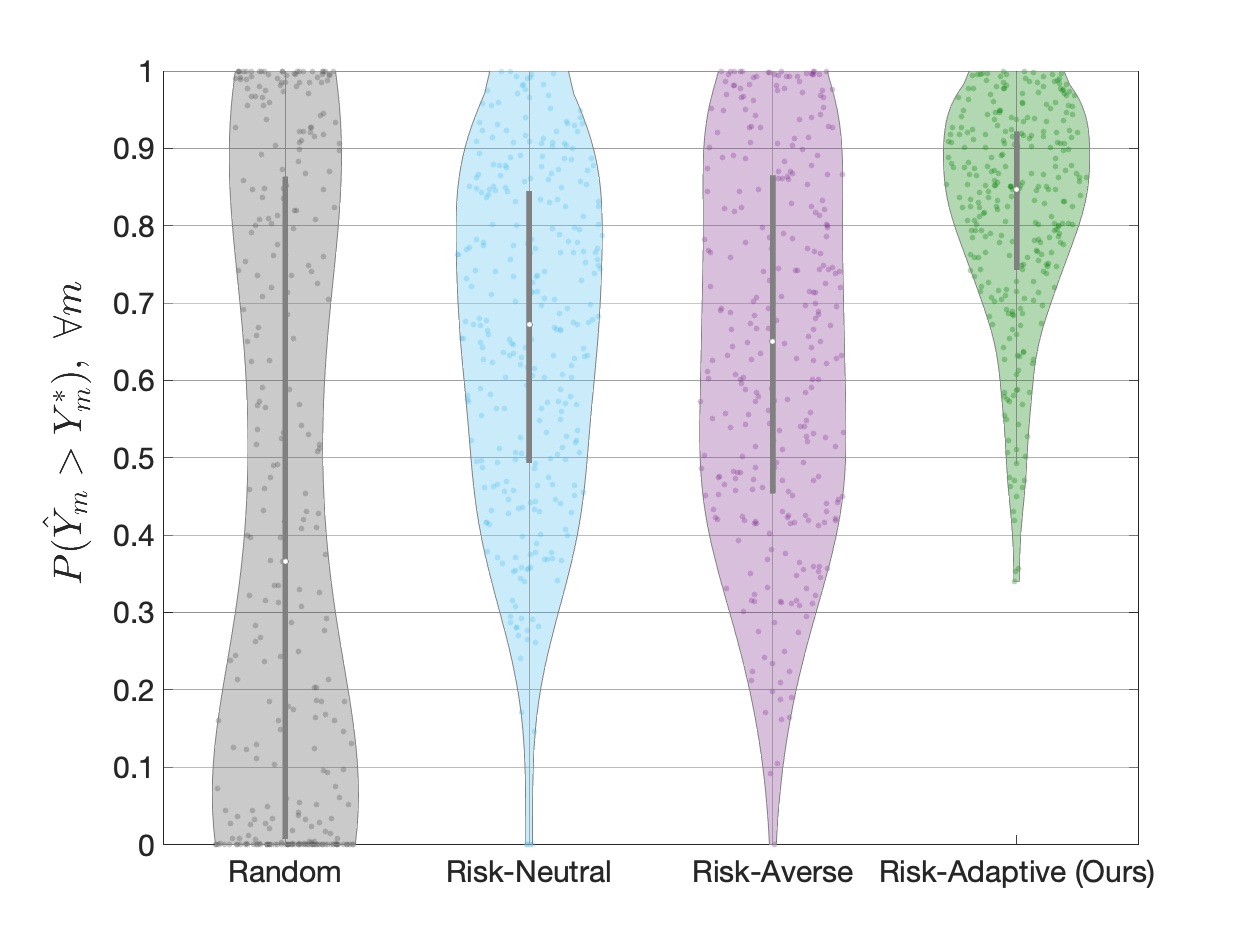}
    \caption{Probability of success for all tasks (300 data points per method).}
    \label{fig:task_probabilities}
\end{figure}

\begin{figure}[ht]
    \centering
    \includegraphics[trim=10 10 10 3,clip, width=1.0\columnwidth]{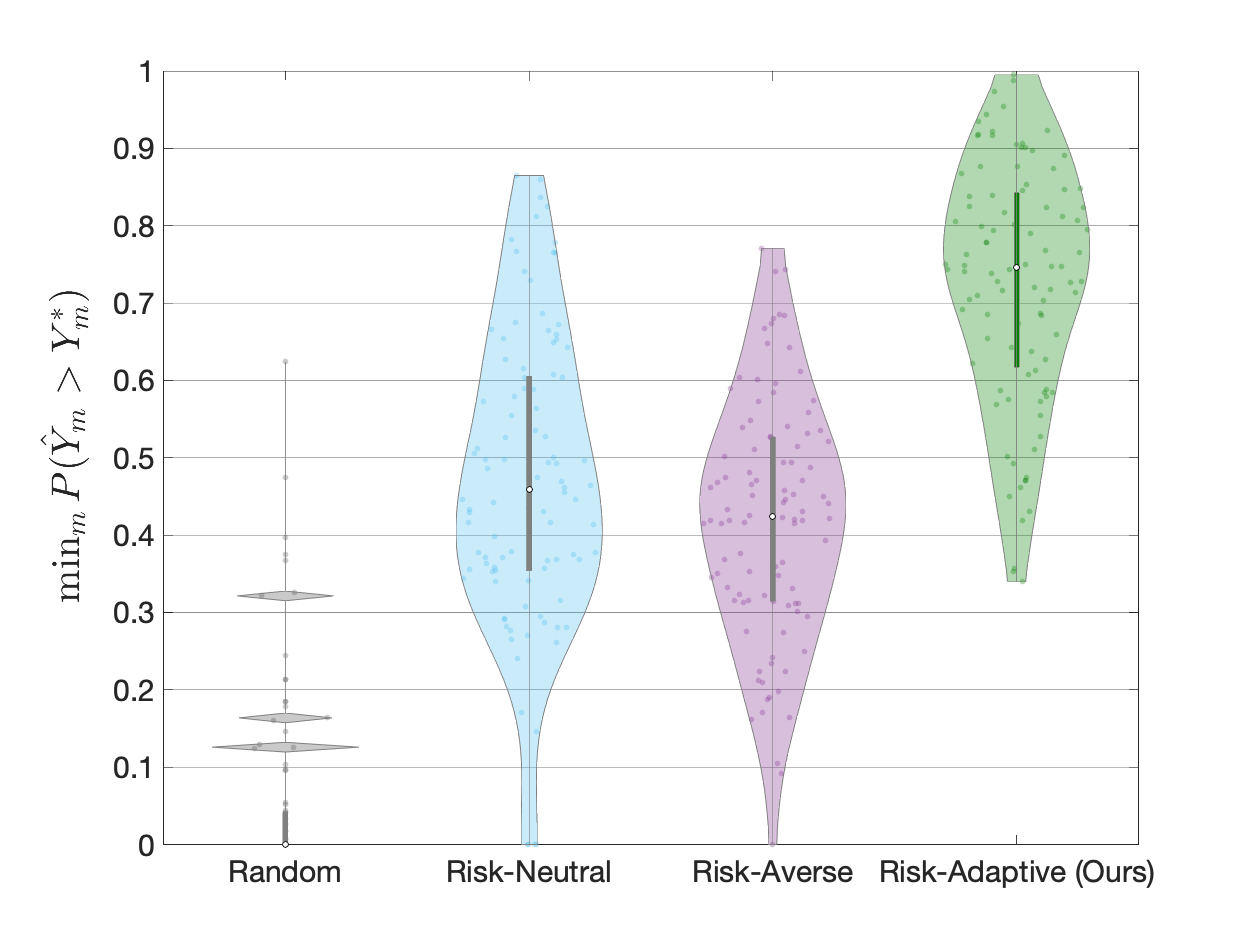}
    \caption{Minimum probability of success computed across tasks (100 data points per method).}
    \label{fig:min_task_probabilities}
\end{figure}

For each run, we measure the performance of each algorithm by computing the success probability for each task, given by $P(\hat{Y}_m > Y^*_m), \forall m=1,2,3$ where $\hat{Y}_m$ denotes the aggregated traits achieved by the candidate algorithm for the $m$th task. Given that the distributions are Gaussian, this metric measures the actual probability of satisfying the task requirements when utilizing a particular allocation rather than providing an approximated rate of success based on Monte Carlo-based simulations.
We report both i) the individual task success probabilities for all the tasks, and ii) the minimum task success probability (computed over the tasks) in Figs. \ref{fig:task_probabilities} and \ref{fig:min_task_probabilities}, respectively. 

From Figs. \ref{fig:task_probabilities} and \ref{fig:min_task_probabilities}, we can see that our risk-adaptive method generally outperformed all the baselines in fulfilling the trait requirement probabilities. This is due to the fact that our approach adaptively chooses between risk-averse and risk-seeking behavior based on the particular allocation problem. Further, thanks to the max-min optimization, our approach ensures that all the chances of success for all tasks are jointly improved. This claim is supported by the considerably lower variance in task success probability across all tasks achieved by our risk-adaptive approach (see Fig. \ref{fig:task_probabilities}).

We observed that the random baseline exhibited the largest variance in individual task success probabilities. This is because the random baseline is more likely to unevenly assign the robots to tasks such that the requirements associated with a subset of the tasks are fulfilled with near-certainty. But, this usually comes at the cost of failing to meet the requirements of the rest of the tasks with near-certainty. From Fig. \ref{fig:min_task_probabilities}, we can see that the random baseline's minimum task success probability per trial was near zero for several instances. 

When looking at the aggregate performance across all 100 runs, we find that the risk-neutral and risk-averse baselines performed similarly to each other. However, for any given problem instance, these two baselines may not necessarily perform similarly. This is due to fact that while avoiding risk might be very helpful in some situations, it might be too conservative in others. Further, the performances of these two baselines are influenced by factors, such as the variances of the trait distributions, and the regularization coefficient $\lambda$. 
However, given the adaptive nature of our approach, it always performs similarly to or better than the baselines (in terms of $\min_m P(Y_m > Y^*)$) for any given instance of the problem. 

In summary, it is evident that our risk-adaptive approach has considerably higher chances of satisfying task requirements compared to approaches that either ignore or avoid risk all together. These observations are to be expected given that our risk-adaptive approach explicitly maximizes the chances of satisfying trait requirements. As explained in \ref{sec:rationale}, this incentivizes the algorithm to adaptively switch between its risk-averse and risk-seeking regimes. 

\subsection{Robotarium Simulations}

\begin{figure}[!h]
    \centering
    \includegraphics[trim=10 10 10 3,clip, width=1.0\columnwidth]{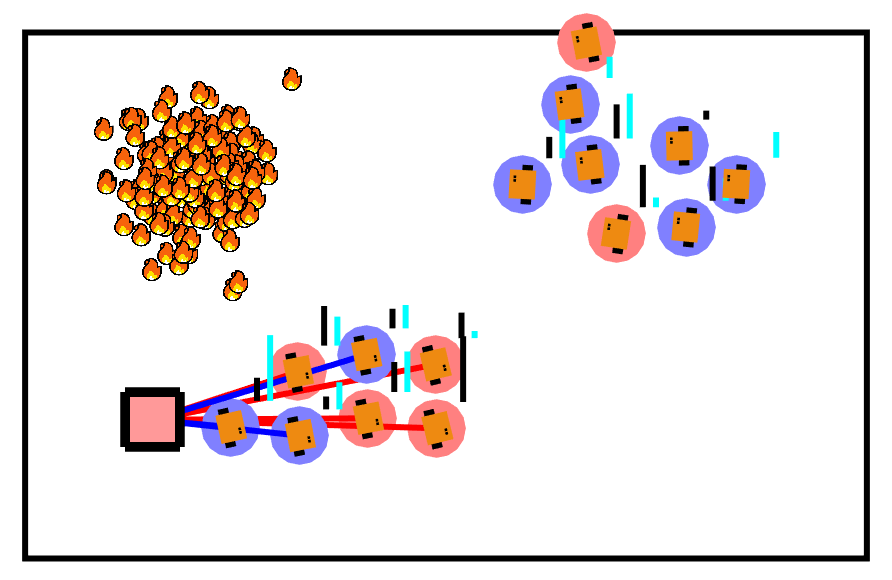}
    \caption{A snapshot of the simulated emergency response task. 
    }
    \label{fig:robot_example}
\end{figure}

In the second round of experiments, we considered a multi-robot scenario to illustrate the benefits of our approach. We developed an emergency response scenario in the \emph{Robotarium} simulator~\cite{wilson2020} in which we sample robot capabilities from specified distributions. Our scenario involved a fire fighting task and a debris removal task (see Fig. \ref{fig:robot_example}). These tasks were to be completed by a heterogeneous team of robots composed of $S=2$ species, each with $U=2$ traits. Species 1 had 6 robots and Species 2 had 9 robots. Each robot had two traits: i) water carrying capacity and ii) payload capacity. The distribution of robot capabilities (in arbitrary units) were modelled using Gaussian distributions with the following parameters:
$$ \bm{\mu}_Q = \begin{bmatrix} 2 & 1\\ 1 & 2 \end{bmatrix}\text{    } \mathrm{Var}_Q = \begin{bmatrix} 0.5 & 1\\ 1 & 0.5 \end{bmatrix}$$

The robots assigned to each task must work together to collectively complete their task.  The debris removal task requires 11 units of strength and the firefighting task requires 14 units of water. More formally, we defined the task requirements matrix as follows:
$$\bm{Y}^* = \begin{bmatrix} 11 & 0 \\ 0 & 14 \end{bmatrix} $$

Note that, if the coalition assigned to the debris removal task do not have the cumulative payload capacity to move the debris, the task would fail. Similarly, if the robots assigned to the firefighting task do not have enough water to douse the flames, the fire burns on.

Using these parameters, we obtained the following allocations computed using each of the methods by solving the corresponding optimization program.
$$X_\text{Ours} = \begin{bmatrix} 6 & 1 \\ 0 & 8 \end{bmatrix} \;\; 
X_\text{R-A} = \begin{bmatrix} 4 & 3 \\ 2 & 6 \end{bmatrix} \;\;
X_\text{R-N} = \begin{bmatrix} 5 & 3 \\ 1 & 6 \end{bmatrix} $$
where R-N and R-A refer to the risk-neutral and risk-averse baselines, respectively. Note that we do not specify the random baseline's assignment matrix, as it would change with every run.

\begin{figure}[ht]
    \centering
    \includegraphics[trim=10 10 10 3,clip, width=0.9\columnwidth]{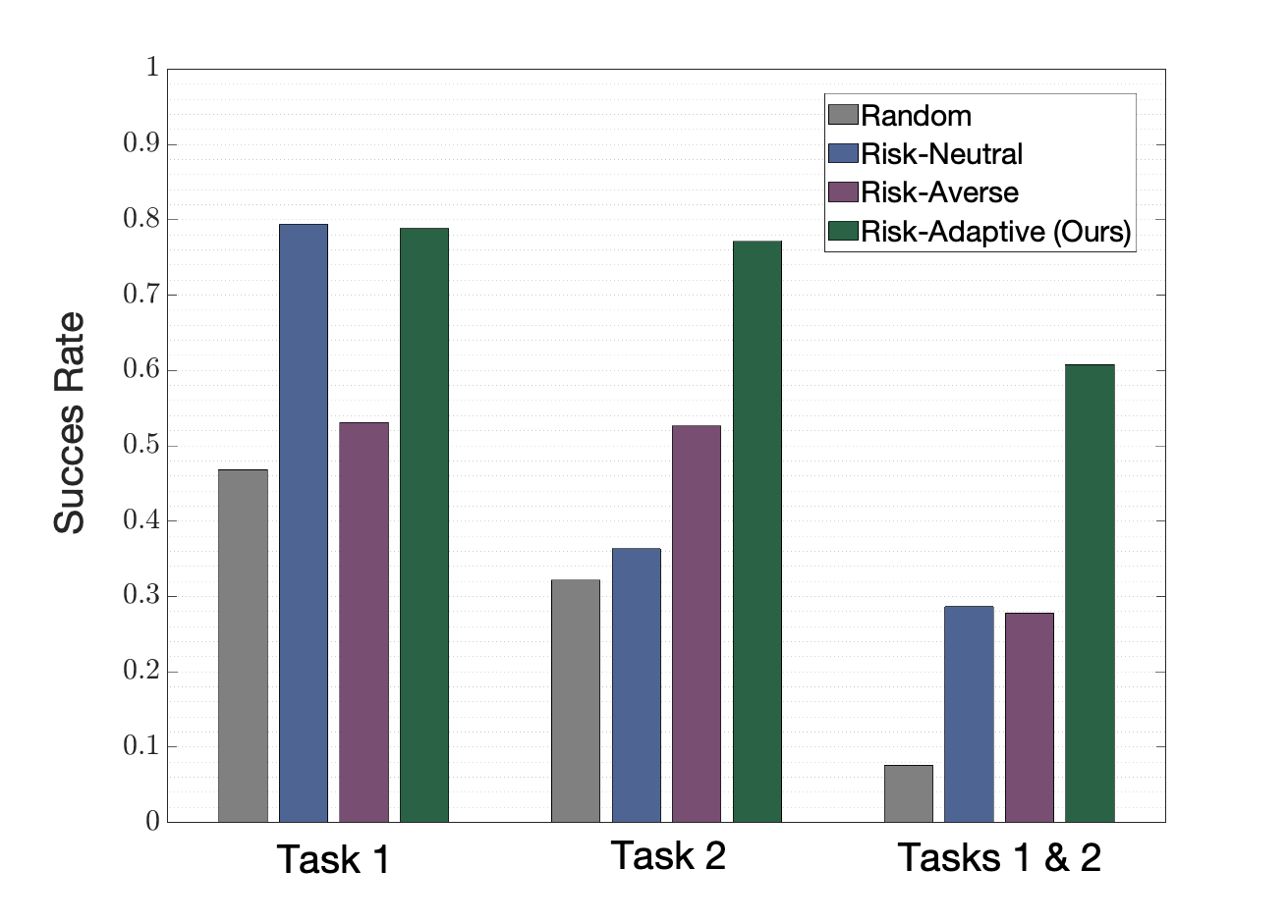}
    \caption{Individual and combined task success rates.}
    \label{fig:robotarium_success}
\end{figure}

To evaluate the approaches on this scenario, we generate $10,000$ instances of the scenario. In each instance, we sampled the robots' traits based on the distribution parameters defined above. We measured the performance of the allocations computed by each approach in terms of task success rates (i.e., no. successful completions / 10,000). We measured both individual task success rates as well as a combined task success rates that required both tasks be completed successfully.
We report these success rates for each approach in Fig. \ref{fig:robotarium_success}. As one would expect, the random baseline performs worse than all other approaches. Further, the risk-neutral and risk-averse approaches outperform each other at different tasks, resulting in similar combined performance. Finally, we can see that our risk-adaptive method successfully completed both tasks at a much higher rate than the baselines as it is more likely to fulfill the corresponding trait requirements.

\section{Conclusion}

We introduced a novel framework for risk-adaptive task allocation that maximizes the probability of satisfying minimum trait requirements instead of maximizing expected pay-off or avoiding worst-case outcomes. Using this framework, we demonstrated that it is necessary to seek risk in order to satisfy requirements when safer options do not meet requirements in expectation. Through numerical simulations and robot experiments, we have shown that our adaptive method indeed results in considerably higher probability of task success. A key limitation of our framework is that we approximately solve our optimization problem using a black-box optimization technique. Further investigation is necessary to leverage any inherent structures of the optimization problem, such as sub-modularity~\cite{prorok_redundant_2019,zhou2020}. 

\addtolength{\textheight}{-12cm}   





\bibliographystyle{IEEEtran}
\bibliography{main}

\end{document}